\documentclass[twocolumn]{article}
\usepackage{booktabs}  % 用于创建漂亮的表格
\usepackage[table,xcdraw]{xcolor}
\usepackage{colortbl}
\usepackage{spconf,amsmath,epsfig}
\usepackage{color}
\usepackage{booktabs}
\usepackage{graphicx}
\usepackage{float}
\usepackage{amsmath}
\usepackage{appendix}
\usepackage{multirow}
\usepackage{threeparttable} %制作三线表格
\usepackage[normalem]{ulem}
\usepackage{lipsum}  % 用于生成示例文本
\usepackage{adjustbox}  % 用于调整表格宽度
\usepackage{tablefootnote}
\usepackage[linesnumbered,ruled,vlined]{algorithm2e}
\useunder{\uline}{\ul}{}
\let\OLDthebibliography\thebibliography
\renewcommand\thebibliography[1]{
  \OLDthebibliography{#1}
  \setlength{\parskip}{0pt}
  \setlength{\itemsep}{0pt plus 0.3ex}
}
\usepackage[backref]{hyperref}
\begin{document}\sloppy
% Example definitions.
% --------------------
%\def\x{{\mathbf x}}https://www.overleaf.com/project/656802f38f5b552bcaec9f1f
%\def\L{{\cal L}}

% Title.
% ------
\title{SSP: A \underline{S}imple and \underline{S}afe automatic \underline{P}rompt Engineering method towards realistic image synthesis on LVM\\}
%
% Single address.
% ---------------
\name{
Weijin Cheng,
Jianzhi Liu,
Jiawen Deng,
Fuji Ren$^{\ast}$ \thanks{*Corresponding author}
}
%Address and e-mail should NOT be added in the submission paper. They should be present only in the camera ready paper. 
\address{
University of Electronic Science and Technology of China (UESTC)\\
    {\tt\small {weijinc0407@outlook.com, renfuji@uestc.edu.cn}}
}

\maketitle
\begin{abstract}
%Recently, text-to-image synthesis (T2I) has great progress through prompt engineering method on Large Vision Models (LVM).
Recently, text-to-image (T2I) synthesis has undergone significant advancements, particularly with the emergence of Large Language Models (LLM) and their enhancement in Large Vision Models (LVM), greatly enhancing the instruction-following capabilities of traditional T2I models. 
Nevertheless, previous methods focus on improving generation quality but introduce unsafe factors into prompts. We explore that appending specific camera descriptions to prompts can enhance safety performance. 
Consequently, we propose a simple and safe prompt engineering method (SSP) to improve image generation quality by providing optimal camera descriptions. 
Specifically, we create a dataset from multi-datasets as original prompts. To select the optimal camera, we design an optimal camera matching approach and implement a classifier for original prompts capable of automatically matching. Appending camera descriptions to original prompts generates optimized prompts for further LVM image generation. 
Experiments demonstrate that SSP improves semantic consistency by an average of 16\% compared to others and safety metrics by 48.9\%. 
\end{abstract}
\begin{figure}[t]
	\centering
	\includegraphics[width=1\columnwidth]{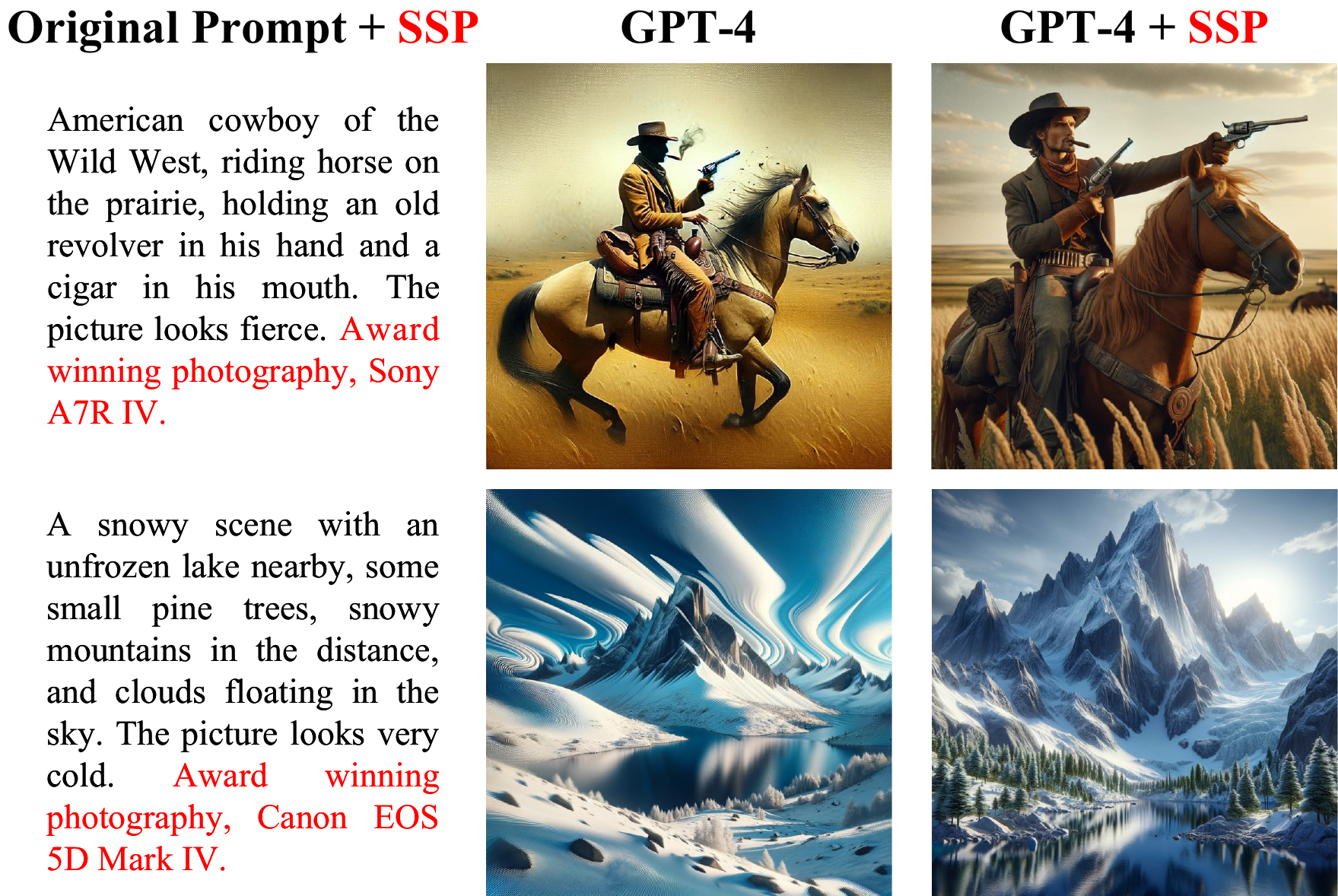}
	\caption{Demonstration of some visual results comparison in GPT-4 image synthesis. SSP can automatically add the optimal camera prompt (red marked) to the original prompt (black marked), which greatly ensures safety (an average of \textbf{48.9\%} improvement compared to other baselines) and prompt consistency (\textbf{16\%} improvement) to meet better image quality.}
 % Demonstration of some visual results comparison in GPT-4 image synthesis. GPT-4 utilizes the original prompt to generate images, resulting in lower image quality. While optimized generation by ChatGPT(ChatGPT opt) can potentially enhance quality, it may introduce semantic or stylistic alterations. SSP offers an automatic enhancement to GPT generation capabilities without requiring GPT fine-tuning and more safety, resulting in improved performance.}
	\label{banner}
\end{figure}

\section{Introduction}
%Text-to-image (T2I) synthesis aims to generate high-quality and realistic images aligned with text inputs.
%the expanding application scope of AI Generated Content (AIGC) has led researchers to shift their focus from Large Language Models (LLMs) to Large Vision Models (LVM)~\cite{bai2023sequential}. 
%Notably, significant progress has been made in text-to-image synthesis (T2I).
%In the early stages, researchers used Generative Adversarial Networks (GANs) for T2I~\cite{reed2016generative}. It later evolved into two main directions: a) diffusion models, including DALL-E 2 \cite{ramesh2022hierarchical}, Imagen \cite{saharia2022photorealistic}, Stable Diffusion \cite{rombach2022high}; and b) autoregressive models, such as CogView \cite{DingYHZZYLZSYT2021CogView}, CogView2 ~\cite{DingZHT2022CogView2}, DALL-E \cite{RameshPGGVRCS2021DALLE}.
%These models can improve the generation quality of T2I. However, the quality of generated images depends on input texts. Models such as DALL-E 2~\cite{ramesh2022hierarchical} and Stable Diffusion~\cite{rombach2022high} exhibit limited text comprehension, capturing only basic vocabulary, and their understanding of complex texts is poor.

Recently, some text-to-image (T2I) synthesis methods, such as ~\cite{nichol2021glide, DingYHZZYLZSYT2021CogView, RombachBLEO22StableDiffusion, DingZHT2022CogView2, SahariaCSLWDGLA22imagen} have undergone significant advancements, particularly with the emergence of Large Language Models (LLM) and their enhancement in Large Vision Models (LVM), greatly enhancing the instruction-following capabilities of traditional T2I models.
%Recently, with the development of the large models~\cite{brown2020language}, text-to-image (T2I) synthesis~\cite{nichol2021glide} researchers have started focusing on Large Vision Models (LVM)~\cite{bai2023sequential}.
%T2I continues to make progress on LVM. 
However there is unable to enhance the performance of LVM through fine-tuning or retraining, researchers have begun searching for methods that can boost the generation performance without the additional training process. The key to improving the generation quality of T2I models lies in designing effective prompts~\cite{dong2022dreamartist}. 
LVM exhibits strong text comprehension~\cite{jiang2020can}, which brings a prompt engineering method~\cite{gao2020making} for better meet specific image generation requirements~\cite{oppenlaender2023taxonomy}.
%to guide LVM in generating high-quality images~\cite{dong2022dreamartist}.

Nevertheless, previous prompt engineering methods involved introducing random additional vocabulary into prompts to generate high-quality prompts. Specifically, BeautifulPrompt~\cite{cao2023beautifulprompt} trains model using both low-quality and high-quality prompt pairs to enhances the quality of the generated images through Reinforcement Learning(RL). %However, the training process of RL is complex, and randomly adding vocabulary leads to changes in the original semantics, partial loss of content, or the introduction of additional unnecessary information. 
% The refinement of the model through Reinforcement Learning is employed in the fine-tuning process, enhancing the overall quality of the generated images.
%NeuroPrompts~\cite{rosenman2023neuroprompts} designs an interactive application that allows the model to automatically optimize prompts, similar to professional human prompt engineers.
BestPrompt~\cite{pavlichenko2023best} selects and compares different candidate keyword sets using genetic algorithms to generate more aesthetically pleasing images.
%employs genetic algorithms to select candidate keywords %from the most popular ones. It then compares images generated by different keyword sets to choose the prompt that generates more aesthetically pleasing images.
%LLM BLUEPRINT~\cite{gani2023llm} iteratively generates scene objects to enhance the understanding of lengthy textual content and facilitate accurate image generation.
PromptMagician~\cite{feng2023promptmagician} is trained using a substantial amount of high-quality prompts, aiming to generate optimal prompts for Stable Diffusion.
However, above approaches that involve the process of randomly adding vocabulary are uncontrolled. This randomness may change the original semantics and introduce unsafe factors, raising safety concerns~\cite{ba2023surrogateprompt}.
%partial loss of content, or the introduction of additional unnecessary information. In some cases, it may even result in the introduction of unsafe factors, raising safety concerns.
% 此处如果能举出例子会比较好一些，mark一下，可以不做，最后有时间可以加一下

To address these challenges, we observe GPT and high-quality image datasets, noting three key points: a) Real-world images must be captured by a camera for computer recognition and storage. Different cameras are dedicated to optimizing different shooting themes (e.g., vibrant color cameras for natural landscapes). b) LVM training data is complex, with some containing camera models; different camera prompts change the image generation effect. c) Appending specific camera descriptions to prompts can prevent the introduction of unsafe factors, enhancing safety performance.

%To address the challenges, we first observed GPT and high-quality image datasets, and came to two observations: a) In order for real-world images to be recognized and saved by computers, they must be captured by a camera. Different cameras are dedicated to optimizing different shooting subjects (for example, cameras with bright colors are more suitable for expressing natural landscapes). b) The training data of LLM has complex components, and some of the data contains camera parameters. Different camera prompts will change the generation effect. 
%(e.g., MSCOCO~\cite{lin2014microsoft}, ImageNet~\cite{deng2009imagenet}, and DiffusionDB~\cite{wang2022diffusiondb})
Therefore, we introduce SSP, a simple and safe prompt engineering method designed to improve image generation quality and avoid introducing unsafe factors by providing camera descriptions. 
Specifically, we first create a dataset sourced from various public text-only or text-image pairs datasets, which is then summarized and filtered by GPT-4~\cite{openai2023gpt4}. Subsequently, we design an optimal camera matching approach by fine-tuning the BERT model~\cite{devlin2018bert} with this dataset, serving as a classifier for original prompts capable of automatic matching. Optimized prompts are obtained by appending the optimal camera descriptions to the originals. We then input these optimized prompts into GPT-4 to generate images. 
%Extensive experiments demonstrate that our method improves image generation quality, achieving both authenticity and safe performance while ensuring alignment with prompts. 
A comparison of prompts and generating images are shown in Fig.~\ref{banner}.

In this study, our main contributions can be summarized as follows: 
\begin{itemize}
  \item We release a new dataset for image generation prompt optimization, suitable for visual prompt optimization tasks.
  \item We introduce SSP, a novel method designed to improve image generation quality by providing optimal camera descriptions without altering the original content or introducing unsafe factors.
  \item Extensive experiments demonstrate the superior performance of SSP compared to two robust baselines. SSP shows an average improvement of 16\% in prompt consistency and 5\% in text-image alignment compared to the baselines, with a 48.9\% increase in safety metrics. 
  \item Experiments with text feature analysis have proved that prompt engineering in LVM can change the data distribution of prompt, so as to achieve better generation results. This conclusion may inspire other strategies for prompt-driven on large model optimization.
\end{itemize}

\begin{figure*}[t]
	\centering
	\includegraphics[width=0.93\textwidth]{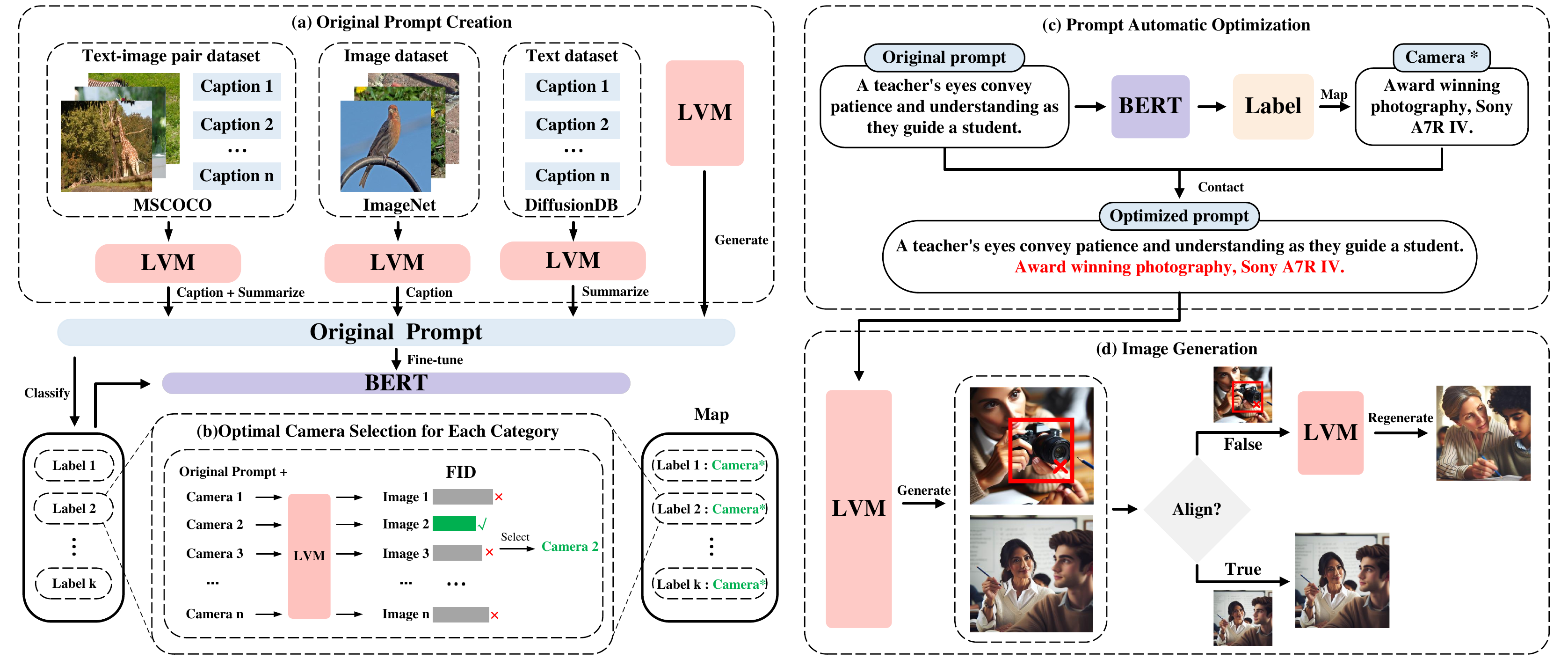}
	\caption{Overview of our proposed SSP. In (a), we use GPT-4 to caption images and summarize captions from multi-datasets, or generate captions directly as original prompts. In (b), we manually classify original prompts based on different shooting themes and select optimal camera for each category by calculating and selecting the lowest FID. Additionally, we fine-tune the BERT model to automatically match optimal camera descriptions for the original prompts. In (c), we input the original prompts into the BERT model to automatically match the optimal camera descriptions, contacting them to generate the optimized prompts. In (d), we input optimized prompts into GPT-4 and output generated images if they align; otherwise, regenerate.}
	\label{Overview}
\end{figure*}

\section{Dataset}
\subsection{Collection of Prompt}
To create original prompts, we firstly use three publicly available datasets as original data sources: MSCOCO~\cite{lin2014microsoft}, ImageNet~\cite{deng2009imagenet}, and DiffusionDB~\cite{wang2022diffusiondb}. Specifically:
a) MSCOCO comprises images paired with corresponding captions, forming text-image pairs. Firstly, we use GPT-4 to caption the images. Subsequently, GPT-4 summarizes both the original and generated captions.
b) ImageNet contains an extensive collection of real images. We extract a subset of these images and employ GPT-4 for caption generation.
c) DiffusionDB contains captions of images. We use GPT-4 to summarize these  captions and generate new captions.
d) Additionally, as GPT-4 can comprehend user input, we prompt it to generate image descriptions directly.
Through above methods, we obtain a candidate set of the original prompts.
Next, we use GPT-4 to filter the candidate set by eliminating texts that are difficult for it to comprehend, and then generate the final set of original prompts. The original prompt collection process is shown in the Fig.~\ref{Overview} (a).

%To create high-quality original prompts, we employ ChatGPT for captioning and summarizing images, or directly generate image descriptions from MSCOCO, ImageNet, and DiffusionDB datasets. We filter out low-quality texts using ChatGPT, resulting in the generation of original prompts.
%The detailed process of dataset collection is provided in the appendix.

\subsection{Optimal Camera Selection}
%We consider diverse cameras for different image genres, selecting common shooting themes and prevalent cameras in the market. Augmenting the original prompt with distinct camera descriptions, we use two metrics to assess the suitability of images generated by GPT-4.

Considering that distinct cameras excel in capturing different image themes, we first choose common shooting themes and prevalent cameras, along with their respective models in the market. We then manually classify the original prompts based on these shooting themes. To select the optimal camera for each category, we add distinct camera descriptions into the original prompts for each category. Subsequently, we input the modified prompts into GPT-4 to generate corresponding images. Each category generates images corresponding to modified prompts containing different camera descriptions. 
%Subsequently, for each image category, distinct camera descriptions were introduced after the original prompt. Both the original prompt and the modified prompt were then input into GPT-4 to generate diverse images corresponding to different input.

To select optimal camera for each category based on the generated images, we introduce two metrics, namely FID~\cite{heusel2017gans} and CLIP Score~\cite{radford2021learning}. FID measures the proximity between generated images and real images, while CLIP Score evaluates consistency between images and prompt. Due to the limited addition of vocabulary resulting in only slight deviation from the original semantics, our primary focus is on ensuring the authenticity of the generated images while preserving the original semantics.
As shown in Fig.~\ref{Overview} (b), we calculate FID of generated images by comparing the distribution difference with real images and select the lowest values for each category. Simultaneously, we ensure minimal CLIP Score decrease. The camera corresponding to the selected values is considered the optimal choice for that category. This process is repeated across all categories to select the optimal camera.
An example of selecting the optimal camera for the portrait category can be found in the appendix.

%Considering that distinct cameras excel in capturing different image genres, we select prevalent shooting themes and identify commonly used cameras along with their respective models in the current market. For each image category, we augment the original prompt with distinct camera descriptions and use two metrics to test the images generated by GPT-4 to decide their suitability.
% generate images from it. resulting in an optimized prompt. Subsequently, both the original prompts and optimized prompts are separately input into GPT-4 to generate the corresponding images.  To determine the most suitable prompt for each category effectively, we introduce two key metrics, namely FID and CLIP Score. we identify the optimal prompt for each image category by experimenting. 

\subsection{Statistics}
We finally collect 10k prompts in our dataset. In Table~\ref{dataset}, the average length of original and optimized prompts is 108.8 and 151.3, respectively. Optimized prompts add an average of only 42.5 characters while maintaining 94\% prompt similarity after optimization, indicating high consistency and a strong alignment between the original and optimized prompts. Comparing to Beautifulrompt, average of 115 fewer characters are added, while prompt consistency increased by 32.4\%.

\begin{table}[htbp]
\resizebox{\linewidth}{!}{
\begin{tabular}{@{}cccccc@{}}
\toprule
\textbf{Dataset} & \textbf{Num} & \textbf{ALLP} & \textbf{ALHP} & \textbf{DALP↓}                     & \textbf{PC↑}                       \\ \midrule
BeautifulPrompt  & 143k         & 40.3          & 197.8         & 157.5                                & 0.71                                 \\
SSP(ours)        & 10k          & 108.8         & 151.3         & \textbf{42.5}
& \textbf{0.94}
 \\ \bottomrule
\end{tabular}
}
\caption{Comparison between BeautifulPrompt dataset and SSP dataset. Note that, ALLP, ALHP, DALP and PC denote the average lengths of original and optimized prompts,  the difference in the average length of prompts, and the prompt consistency (we use CLIP text cosine similarity to evaluate it), respectively.}
\label{dataset}
\end{table}

\section{Methods}
\subsection{Optimized Prompt Generation}
To automatically match the optimal camera and generate optimized prompts, we 
first design an optimal camera matching approach as shown in Fig.~\ref{Overview} (c).
Specifically, we fine-tune BERT with the original prompts as input. The model generates corresponding labels, which are then mapped to the respective camera descriptions. In practical applications, BERT functions as a classifier, facilitating automatic matching from original prompts to camera descriptions during this process. 
Finally, these camera descriptions are added to the corresponding original prompts, resulting in the generation of optimized prompts. 
Detailed parameter settings for fine-tuning BERT are provided in the appendix.

\subsection{Image Generation}
To evaluate the image generation quality based on different prompts and validate the effectiveness of our proposed method, we input the original and optimized prompt sets individually into GPT-4, resulting in two corresponding sets of generated images. Specifically, we use \textbf{BERT} as the automatic prompt optimization model, as outlined in Algorithm~\ref{alg:prompt_optimization}, where $P$ represents the original prompts, and $P^*$ represents the optimized prompts. 
By inputting $P$ into our fine-tuned \textbf{BERT} model on our dataset, we obtain a label $L$. Through the \textbf{LabelToText} process, $L$ is mapped to the camera description $Camera^*$ corresponding to $P$. Afterward, $Camera^*$ is added to $P$, resulting in $P^*$. Subsequently, the process \textbf{GPTGenerate} is applied to \(P^*\), generating the image \(I\)(see Algorithm~\ref{alg:prompt_optimization}, line~\ref{line:generated_image}).

\begin{algorithm}[t]
\caption{Prompt Automatic Optimization and Image Generation. \textbf{BERT} is our automatic prompt optimization model. \textbf{GPTQuery} is a function to ask GPT whether the generated image aligns with the prompt. \textbf{GPTRegenerate} is using a special prompt to regenerate image.}
\label{alg:prompt_optimization}
\KwData{\(P,P^*\) (string) - original and optimized prompt}
\KwResult{\(I\) (image) - generated or regenerated image}

\(L = \textbf{BERT}(P)\) \;
\label{line:optimized_prompt}

\(Camera^* = \text{{LabelToText}}(L)\) \;
\(P^* = (P, \text{{Camera}}^*)\) \;

\(I = \textbf{GPTGenerate}(P^*)\) \;
\label{line:generated_image}

\If{\textbf{GPTQuery}(I,P)}{
    \(I = \textbf{GPTRegenerate}(I, P^*)\) \;    
}
\end{algorithm}

\begin{figure*}[t]
	\centering
	\includegraphics[width=0.92\textwidth]{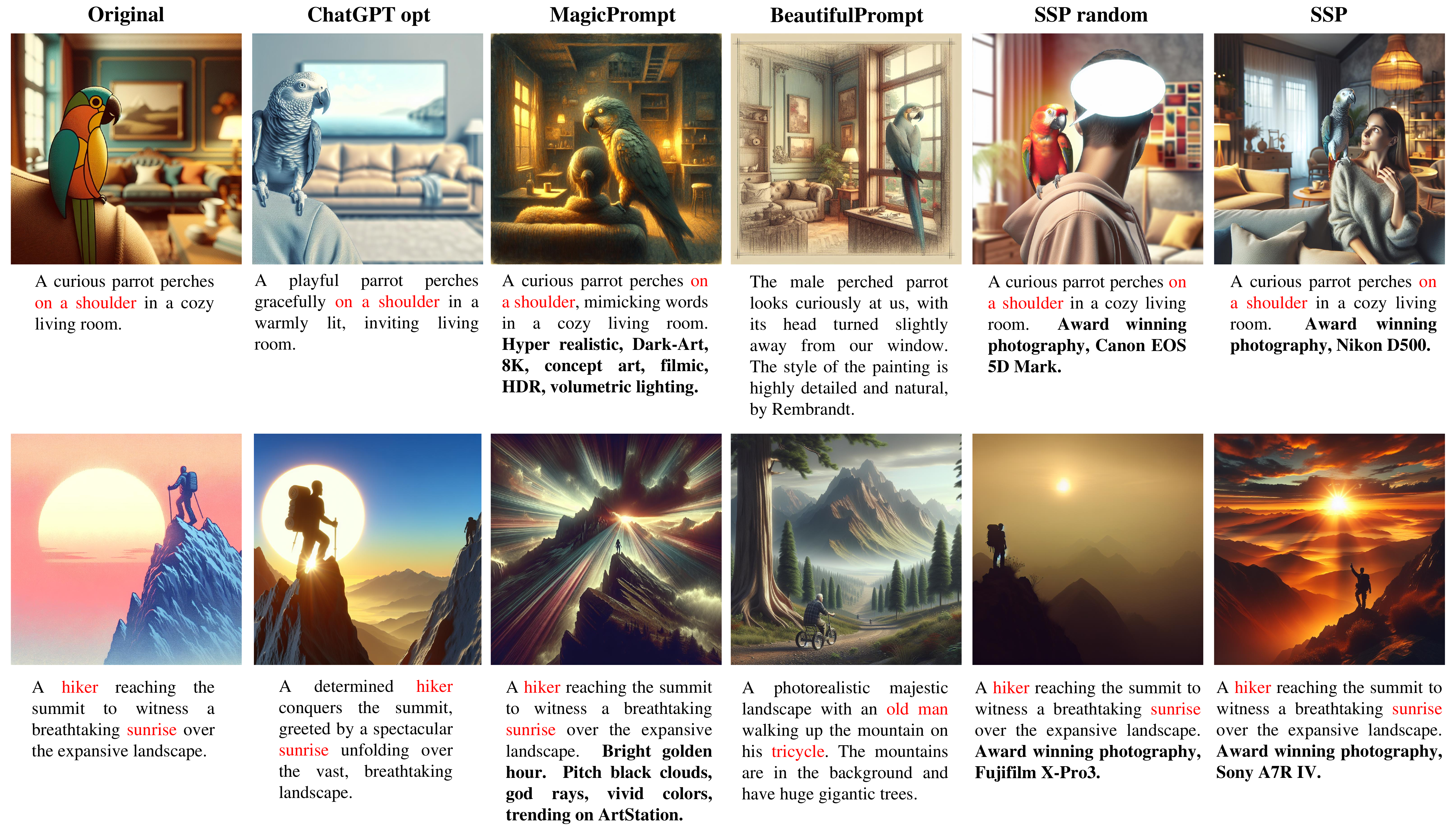}
	\caption{Comparison of prompts and generated images using different methods. Red font denotes errors or omissions from other methods. Bold text indicates additions directly after the original prompt without altering it.}
	\label{visual}
\end{figure*}

Despite the powerful text encoding capabilities and semantic understanding of GPT-4, there are limitations in accurately capturing human intent. For example, in Fig.~\ref{Overview} (d), GPT-4 incorrectly interprets ``camera shooting'' as ``generating camera'', deviating from the original semantics and failing to align with the original prompts.
%the generated image at the top contradicts the intended meaning of the input text description, which specifies that the image is captured by a designated camera. The model directly generates an image of a camera, deviating from the originally intended interpretation. 
Therefore, ensuring alignment between the generated image content and the given prompt is a crucial consideration when using prompts as model inputs.

To address misunderstanding and ensure alignment between the generated content and human intent, 
we use \(\textbf{GPTQuery}\) to query GPT-4 about whether the generated image aligns with the prompt, as depicted in Algorithm~\ref{alg:prompt_optimization}. If output contents are deviated from the original semantics, the \(\textbf{GPTRegenerate}\) process is initiated. This process involves using a specific prompt to instruct GPT-4 to regenerate the image using the current image \(I\) and the optimized prompt \(P^*\) as inputs, subsequently treated as the final output. Conversely, if \(I\) are aligned with the \(P\), the image \(I\) is directly output. Through this methodology, we aim to enhance the accuracy of generated images, ensuring the generation of images aligned with human intent. The entire reasoning process is illustrated in Fig.~\ref{Overview} (d).

\section{Experiment}
\subsection{Settings}
% 这个不属于相关工作 属于评价标准
\textbf{Metrics} To comprehensively assess the quality of the generated images and optimized prompts, several metrics are introduced. 
a) \textbf{CLIP Score} quantifies the semantic consistency between prompt-image pairs and among prompts by comparing their feature vectors. Due to the limited maximum text length that CLIP can handle, we segment long texts into equal-length sequences for computation and calculate the average as the final result.
b) \textbf{FID} calculates the distribution divergence between generated and real images, indicating their proximity to the real world.
c) \textbf{User study} involves multiple assessors subjectively evaluating generated images and providing intuitive feedback on the image quality and text consistency. 
d) \textbf{Detoxify}~\cite{Detoxify} evaluates text toxicity in various dimensions, including toxicity, severe toxicity, obscenity, threat, insult, and identity attack. Scores range from 0 to 1, with lower scores indicating safer content. The final result is the average across these dimensions.

\textbf{Baselines} We introduce three robust baselines: ChatGPT\cite{openai2023chatgpt}, MagicPrompt~\cite{feng2023promptmagician}, and BeautifulPrompt. ChatGPT has robust text comprehension capabilities, allowing it to serve as an expert prompt engineer. MagicPrompt is trained on high-quality prompts for automatic optimization. BeautifulPrompt enhances prompts using supervised fine-tuning and Reinforcement Learning for higher image quality.

\subsection{Overall Results}
\subsubsection{Qualitative Analysis}
To compare optimized prompts and the visual effect of generated images by different methods, we input optimized prompts generated by various approaches into GPT-4 to generate images. As shown in the fourth image of the first row in Fig.~\ref{visual}, it displays two sets of images from different methods, illustrating that our approach produces realistic and aesthetically appealing results. The generated content aligns well with the input prompt compared to other methods. 
For instance, the original prompt mentions ``a parrot on a shoulder'',while BeautifulPrompt misses ``shoulder'', depicting only a parrot in the image and directly altering the theme of the prompt.
%We optimize the original prompts in our dataset using various approaches. These optimized prompts are then used as inputs for GPT-4 to generate images. Figure~\ref{visual} shows two sets of images from different methods, and it's clear from the visuals that our method produces realistic and aesthetically appealing images. Additionally, the generated content aligns well with the input text compare to other methods. For example, the original prompt mentions the violin, while the BeautifulPrompt optimization has altered it to the piano, deviating from the original mention of the violin, thus directly changing the central theme of the text.

\subsubsection{Quantitative Analysis}
In Table~\ref{result}, our approach outperforms other baselines across diverse metrics.
To validate the authenticity of generated images, we conduct a comparative FID against various methods.
%where lower FID value indicates closer proximity to the real world. 
Our method achieves the minimum FID, indicating our generated images are closer in proximity to real images. 
To assess the alignment between generated images and prompt, we evaluate CLIP Score among different methods. 
While the introduction of content to the prompt leads to a slight reduction in CLIP Score, our approach experiences the slightest decrement, which is only 0.05 from the Original. This result illustrates the maximal alignment of generated images with the prompt.
To assess the visual effects of different methods, we compare human preferences based on both image quality and textual consistency with other approaches. Our approach's winning rate, depicted in Fig.~\ref{win}, exceeds 65\%. The detailed design and scoring criteria for the user study can be found in the appendix.

\begin{table}[h!] %表格的浮动环境
\resizebox{\linewidth}{!}{
\begin{tabular}{@{}cccc@{}}
\toprule%表头直线
\textbf{Method}  & \textbf{FID↓} & \textbf{Alignment↑} & \textbf{User Study↑} \\ \midrule%表中直线
Original          & 1.645  & \textbf{30.93}            & 4.87                     \\
ChatGPT opt       & 1.670           & 30.85            & 5.34                    \\
MagicPrompt       & {\ul 1.604}         & 29.24             & 5.29                    \\
BeautifulPrompt       & 1.620           & 29.61              & {\ul 5.96}                    \\
\begin{tabular}[c]{@{}c@{}}
SSP random  \end{tabular} & 1.640          & 29.76             & 5.34                       \\
SSP(ours)           & \textbf{1.596}     & {\ul 30.88}         & \textbf{8.32}                       \\ \bottomrule%表底直线
\end{tabular}
}
\caption{Results on our test set. ``Alignment'' denotes the text-image alignment through CLIP Score.}
\label{result}
\end{table}

\begin{figure}[t]
	\centering
	\includegraphics[width=0.75\columnwidth]{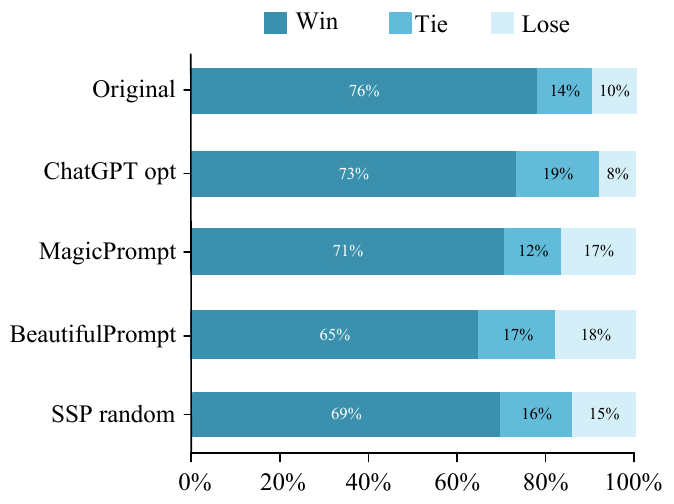}
	\caption{Comparison of human preference evaluation (i.e.,win/lose/tie rates of SSP against other methods).}
	\label{win}
\end{figure}

\subsubsection{Unsafe Factor Detection}
To validate whether the process of optimizing prompts introduces any unsafe factors, we use Detoxify for text toxicity analysis. As shown in the Table~\ref{safe}, our method produces the lowest Detoxify scores. Furthermore, our method exhibits the lowest rejection rate on GPT-4 built-in safety checks. This underscores the superior safety of our generated content.% compared to alternatives.

\begin{table}[t] %表格的浮动环境
\resizebox{\linewidth}{!}{
\begin{tabular}{cccccc}
\toprule
\textbf{Method}   & \textbf{Original} & \textbf{ChatGPT} & \textbf{MagicPrompt} & \textbf{BeautifulPrompt} & \textbf{SSP(ours)}      \\ \midrule
Detoxify↓(\(10^{-5}\)) & 49.23    & 45.09   & 59.50       & 60.85           & \textbf{40.42}  \\ 
Reject Rate↓ & 5.84\%    & 3.74\%   & 21.31\%       & 17.92\%           & \textbf{1.20\%}  \\  \bottomrule
\end{tabular}
}
\caption{Comparing the security of prompts generated by different methods. Detoxify represents the toxicity assessment of the generated prompts. Reject Rate signifies the probability of GPT-4 rejecting to generate images.}
\label{safe}
\end{table}

\subsubsection{Prompt Analysis}
To evaluate the quality of optimized prompts, we compare the similarity of prompts optimized by different methods with the original prompts. As shown in the Table~\ref{text}, our method requires the fewest additional characters to generate optimized prompts and exhibits the highest textual similarity to the original prompts. 
%indicating that our optimization process maximally preserves the original semantics.

%Prompts from different optimization methods are shown in Fig.~\ref{visual}. The original mentions a parrot on the shoulder, while BeautifulPrompt omits the shoulder and MagicPrompt adds excessive vocabulary. In contrast, our method minimizes content while preserving the original semantics.
%The prompts generated by various optimization methods are presented in the figure~\ref{prompt}. It's evident that our method introduces the least amount of content, preventing the inclusion of uncontrollable elements and maintaining the original semantics. 

\begin{table}[h]
\centering%居中
\begin{tabular}{llll}
\toprule%表头直线
\textbf{Method}   & \textbf{DALP↓} & \textbf{PC↑}  \\ \midrule%表中直线
MagicPrompt           & 108.3       & 0.91      \\
BeautifulPrompt       & 108.5       & 0.71      \\
SSP random            & 42.7        & 0.93      \\
SSP(ours)             & \textbf{41.1}   & \textbf{0.94}  \\ \bottomrule%表底直线
\end{tabular}
\caption{Comparing prompt consistency through different methods. Note that, DALP and PC denote the difference in the average length of prompts and the prompt consistency (we use CLIP text cosine similarity to evaluate it), respectively.}
\label{text}
\end{table}

\subsubsection{Prompt Text Feature Analysis}%根据老师的思路思考一下要不要改
To assess the impact of our method on prompt text features, we use PCA visualizations to reduce high-dimensional text data to a 2-dimensional representation. 
The detailed PCA computation process is provided in the appendix.
As illustrated in Fig.~\ref{pca},
%In the left graph, the green points exhibit greater concentration than the gray points, indicating that our method produces prompts with more compact text features in the space. In contrast, the right graph indicates a close alignment between the distribution of red points and the gray points, suggesting that prompts optimized by ChatGPT show no significant differences from the original prompts. 
in the left graph, the concentration of green points is higher than that of gray ones, indicating that our method generates prompts with compact text features in the feature space. In contrast, the right graph demonstrates a close alignment between red and gray points, suggesting that optimized prompts by ChatGPT show no significant differences from the original prompts.
This indicates that our method more easily influences the text features of prompts.
As shown in Table~\ref{result}, this influence is positive and contributes to improving the generation effect of the LVM, resulting in the production of realistic and visually appealing images.

\begin{figure}[t]
	\centering
	\includegraphics[width=0.76\columnwidth]{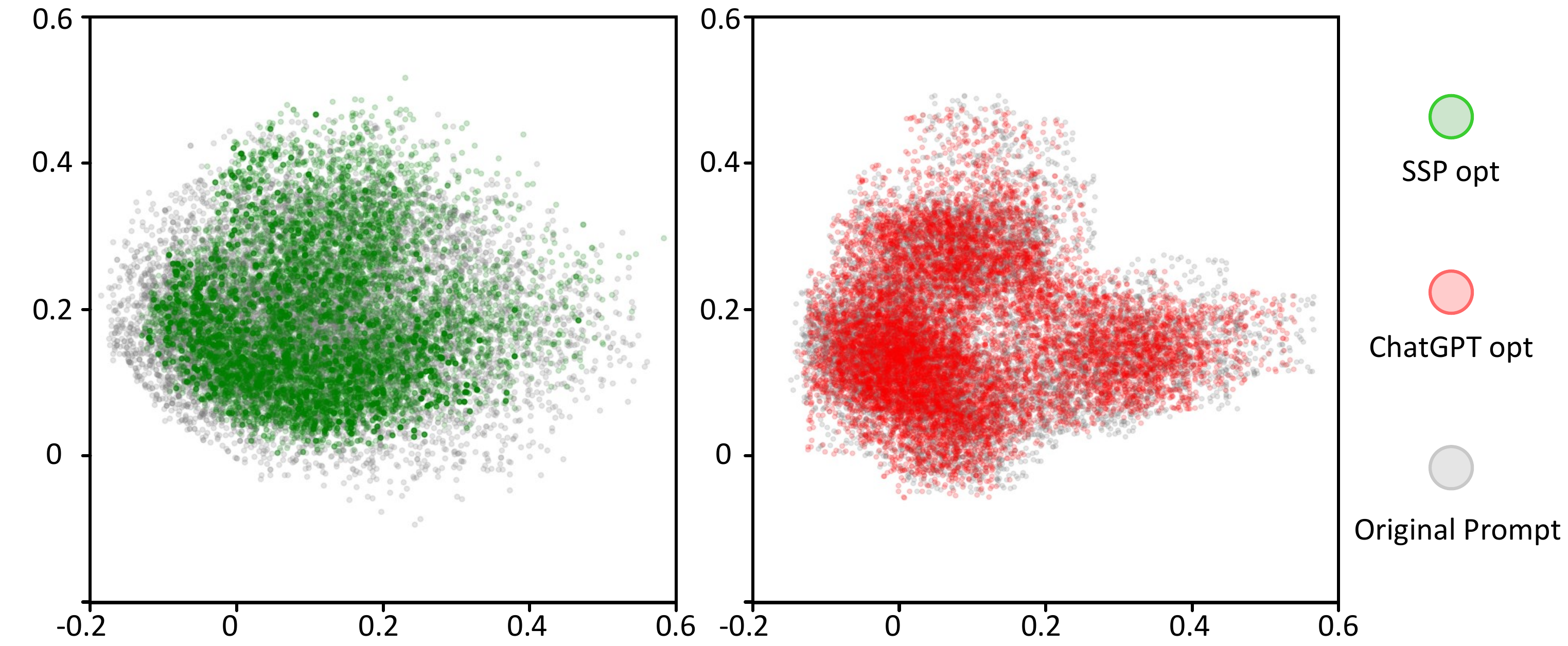}
	\caption{PCA analysis of text feature distribution between original and optimized prompts through different methods. As shown in the legend, different colors represent the encoding results of prompts generated by different models.% On the left, the SSP chart displays green dots for optimized prompts and gray dots for the original prompts. On the right, the ChatGPT chart shows red dots for optimized prompts and gray dots for the original prompts.
 }
	\label{pca}
\end{figure}

\section{Conclusion}
We propose a simple and safe method named SSP to optimize prompts, generating high-quality and aesthetically realistic images using LVM. However, there are certain limitations. a) The validation of authenticity relies solely on FID between real and generated images, lacking dedicated metrics. b) Due to the limited numbers of accessibility of available LVMs, there is a shortage of comparisons with other LVMs. Consequently, future research will prioritize authenticity assessment metrics for images and more universally applicable prompt engineering methods. Additionally, we have only focused on common categories for image generation, future work will explore the versatility of the prompt engineering method.

\appendix

\section{Related works for Text-to-image (T2I) synthesis}
%\subsection{Text-to-image (T2I) synthesis}
% 句子里面不要出现两个and 记得引用GAN的代表作
Text-to-image (T2I) synthesis is a multi-modal task focused on generating aesthetically pleasing and realistic images based on textual descriptions. 
In the early stages, researchers used Generative Adversarial Networks (GANs) for T2I~\cite{reed2016generative}. It later evolved into two main directions: a) diffusion models, including DALL-E 2 \cite{ramesh2022hierarchical}, Imagen \cite{saharia2022photorealistic}, Stable Diffusion \cite{rombach2022high}; and b) autoregressive models, such as CogView \cite{DingYHZZYLZSYT2021CogView}, CogView2 ~\cite{DingZHT2022CogView2}, DALL-E \cite{RameshPGGVRCS2021DALLE}.
With the advent of advanced large-scale models, a significant transformation has occurred in the T2I domain. These models have tremendously enhanced T2I's comprehension of textual input, leading to a surge in its capabilities. Concurrently, this has resulted in an increased reliance on textual descriptions to drive image generation processes. 

\section{optimal camera selection}\label{aboutdataset}

%\subsection{optimal camera select}
To select the optimal camera for prompts in the portrait category, we generate the images from each optimized prompts by GPT-4, and test the indicator.
In Table~\ref{camera}, when the camera is set to Sony A7R IV, the FID value is minimized, indicating the generated images are closest to real images. Notably, the CLIP Score shows a slight decrease, maintaining a high level with only a 0.11 difference from the second-place CLIP Score. Consequently, Sony A7R IV was selected as the optimal camera for this specific image category.

% \usepackage{booktabs}
% \usepackage[normalem]{ulem}
\begin{table}[htbp]
\resizebox{\linewidth}{!}{
\begin{tabular}{@{}cccc@{}}
\toprule
\textbf{Category} & \textbf{Camera}      & \textbf{FID(↓)} & \textbf{CLIP Score(↑)} \\ \midrule
\multirow{11}{*}{Portrait}                  & Original             & 3.003           & 30.17                  \\
          & Canon EOS 90D        & 2.857           & \textbf{31.43}         \\
                  & Sony A7R IV          & \textbf{2.677}  & \textit{30.52}         \\
                  & Fujifilm X-T4        & \textit{2.836}  & 30.50                  \\
                  & Canon EOS 5D Mark IV & 3.019           & 29.94                  \\
                  & Nikon D500           & 2.868           & 30.26                  \\
                  & Sony Alpha 7R IV     & 2.924           & 30.15                  \\
                  & Nikon D6             & {\ul 2.775}     & 30.19                  \\
                  & Sony A7S III         & 2.872           & 29.96                  \\
                  & GoPro HERO9 Black    & 2.865           & {\ul 30.63}            \\
                  & Fujifilm X-Pro3      & 2.899           & 30.39                  \\ \bottomrule
\end{tabular}
}
\caption{An example of the selection of the optimal camera for portrait category. }
\label{camera}
\end{table}

%\subsection{some of dataset text} % 稍微描述一下dataset 的样子，然后举几个例子，什么文本变成什么文本了这样

\section{Fine-tuning Setting Details}
To generate camera descriptions for each original prompt, we fine-tune the BERT model~\cite{devlin2018bert} on our dataset, which is divided into an 8k training set and a 2k testing set. We use a pre-training checkpoint based on the Masked Language Model (MLM) architecture of BERT, and the BERT model has a total of 12 hidden layers. We set the hidden size to 768, the intermediate size to 3072, the number of attention heads to 12, and the number of hidden layers to 12, with a vocabulary size of 30522. All experiments are implemented in PyTorch and run on a single server with NVIDIA Tesla V100 GPUs.

\section{User Study}
To obtain user feedback and assess the effectiveness of our method, we engage 10 assessors for a randomized assessment, meticulously collect evaluation data. Statistical analysis yields average scores for each criterion, furnishing precise quantitative outcomes.
To comprehensively consider image quality and text-visual consistency, we formulate a scoring mechanism:
\[
\text{Score}_i = w_Q \cdot Q_i + w_C \cdot C_i
\]

\begin{figure}[h]
	\centering
	\includegraphics[width=1\linewidth]{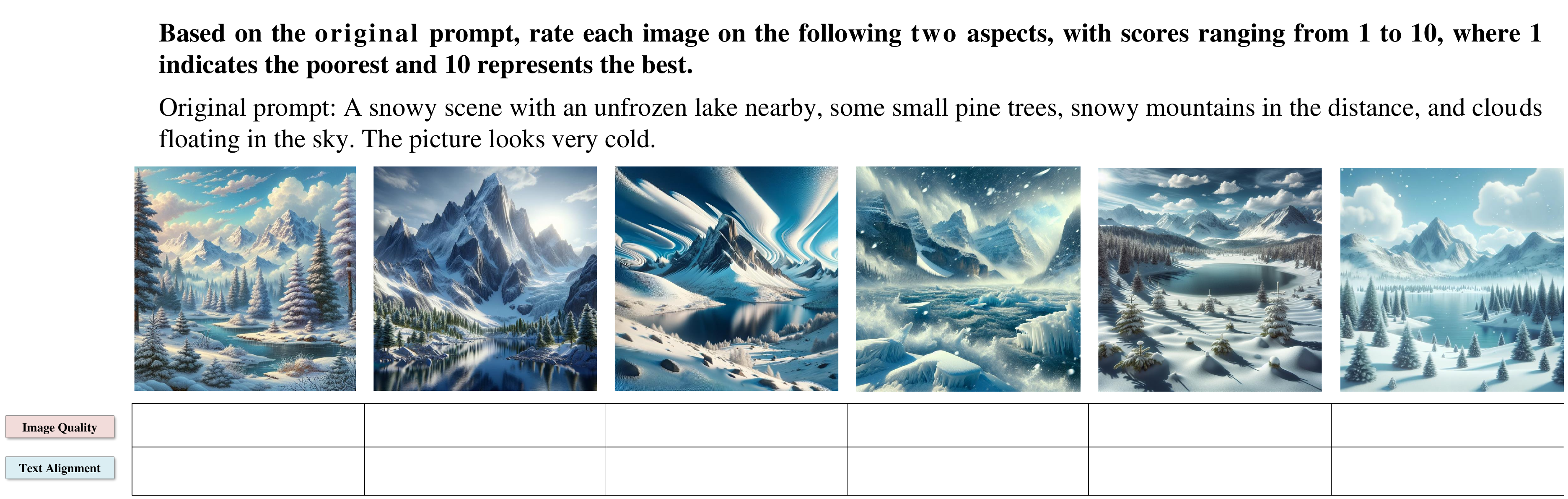}
	\caption{An example of user study scoring interface and criteria for the human evaluation experiment.}
	\label{us}
\end{figure}

Here, \(\text{Score}_i\) represents the overall score for the \(i\)-th image, \(Q_i\) denotes the image quality score, \(C_i\) represents the text-image consistency score, and \(w_Q\) and \(w_C\) are the respective weights. This scoring system provides a comprehensive assessment of various facets in image generation.
The detailed scoring interface is as shown in figure~\ref{us}.

\section{Prompt Text feature Analysis}
To visually analyze the distribution of optimized prompt text features compared to the original prompt using PCA, the CLIP text encoder (\( \text{E} \)) derives features from \( X \):
\[
\text{TF}(X)_{ijk} = \sigma(\text{E}_{ijk}(X))
\]
Resulting in text features of dimensions \( [n, l, d] \). The three-dimensional features are reshaped into \( [n, (l * d)] \) through element-wise multiplication:
\[
\text{RF}(X)_{ij} = \text{TF}(X)_{ijk} \times \text{TF}(X)_{ij(k+1)}
\]
Finally, PCA is applied to the reshaped features (\( \text{RF}(X) \)), projecting onto a two-dimensional plane using the top two eigenvalues.
\[
\text{PCA}(X) = \text{RF}(X) \times \text{EigVec}
\]
Resulting in \( [n, 2] \), encapsulating crucial information for visualization and analysis.

% The PCA visualizations in Figure~\ref{pca} reveal distinct patterns. In the left graph, the green points are more concentrated than the gray points, indicating that our method produces optimized prompts with more compact text features in the space. Conversely, in the right graph, the red points' distribution closely aligns with the gray points, suggesting that ChatGPT's optimized prompts do not exhibit significant differences from the original prompts.
% Our approach positively influences prompt text expression, as evident in the experimental results, shown in Table~\ref{result}. The optimized prompts generated by our method enhance the LVM's ability to create realistic and visually appealing images.

% \begin{figure}[htbp]
% 	\centering
% 	\includegraphics[width=1\columnwidth]{fig/PAC.pdf}
% 	\caption{PCA analysis of text feature distribution between original and optimized prompts through different methods. On the left, the SSP chart displays green dots for optimized prompts and gray dots for the original prompts. On the right, the ChatGPT chart shows red dots for optimized prompts and gray dots for the original prompts.}
% 	\label{pca}
% \end{figure}

\section{More discussing}
\subsection{Why do we publish a dataset for testing rather than use public datasets? }
As the LVM is trained on the open domain, its training process incorporates diverse datasets rather than relying on a specific singular dataset. Consequently, the evaluation of Artificial Intelligence Generative Content (AIGC) extends beyond the confines of a particular dataset, with a significant portion of assessments being conducted through the collection of publicly available datasets or via web searches. This allows LLM to summarize or generate content, thereby simulating the training of large models using a variety of data sources.

To assess the performance of LVM in image generation, we use multiple datasets as our data sources, including MSCOCO, ImageNet, and DiffusionDB. We caption images using GPT-4 across these diverse datasets, summarizing the original captions alongside those generated by the model. Subsequently, we directly generate descriptions through ChatGPT, systematically screening and filtering the text obtained from these datasets to eliminate challenging content specific to LVM. This meticulous process ultimately led to the derivation of the original prompt for our study.

\subsection{Why do we only release a small dataset contains 10k data items? }
The previous approaches employ Reinforcement Learning, requiring extensive dataset to complete the prompt optimization process. In contrast, our approach utilizes a simple BERT classification model, achieving training effectiveness with just 10k data.

\section{More visual results}
\begin{figure*}[t]
	\centering
	\includegraphics[width=\textwidth]{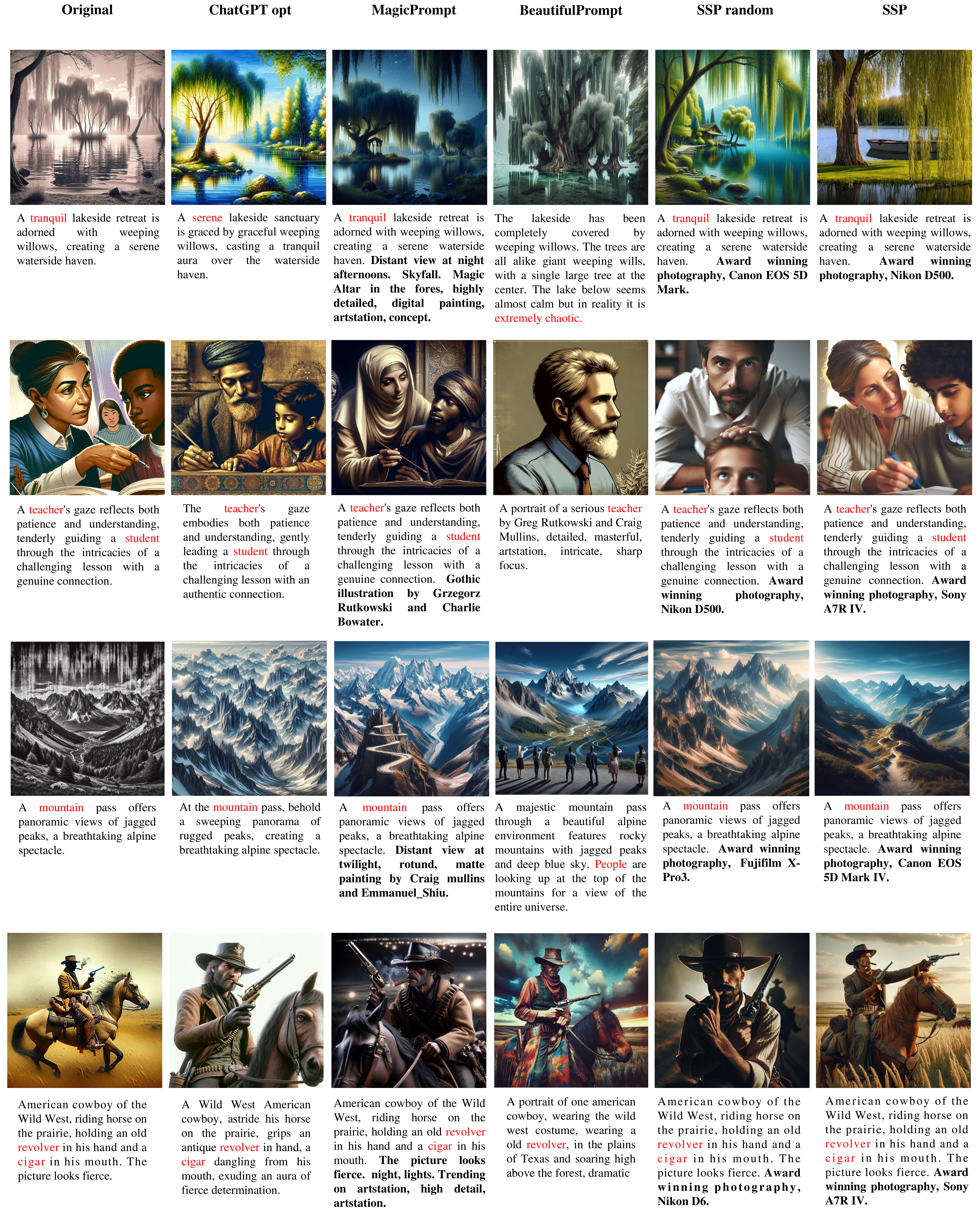}
	\caption{Examples of prompts and images generated by different methods. Red font denotes errors or omissions from other methods. Bold text indicates additions directly after the original prompt without altering it.}
	\label{tu1}
\end{figure*}

\begin{figure*}[t]
	\centering
	\includegraphics[width=\textwidth]{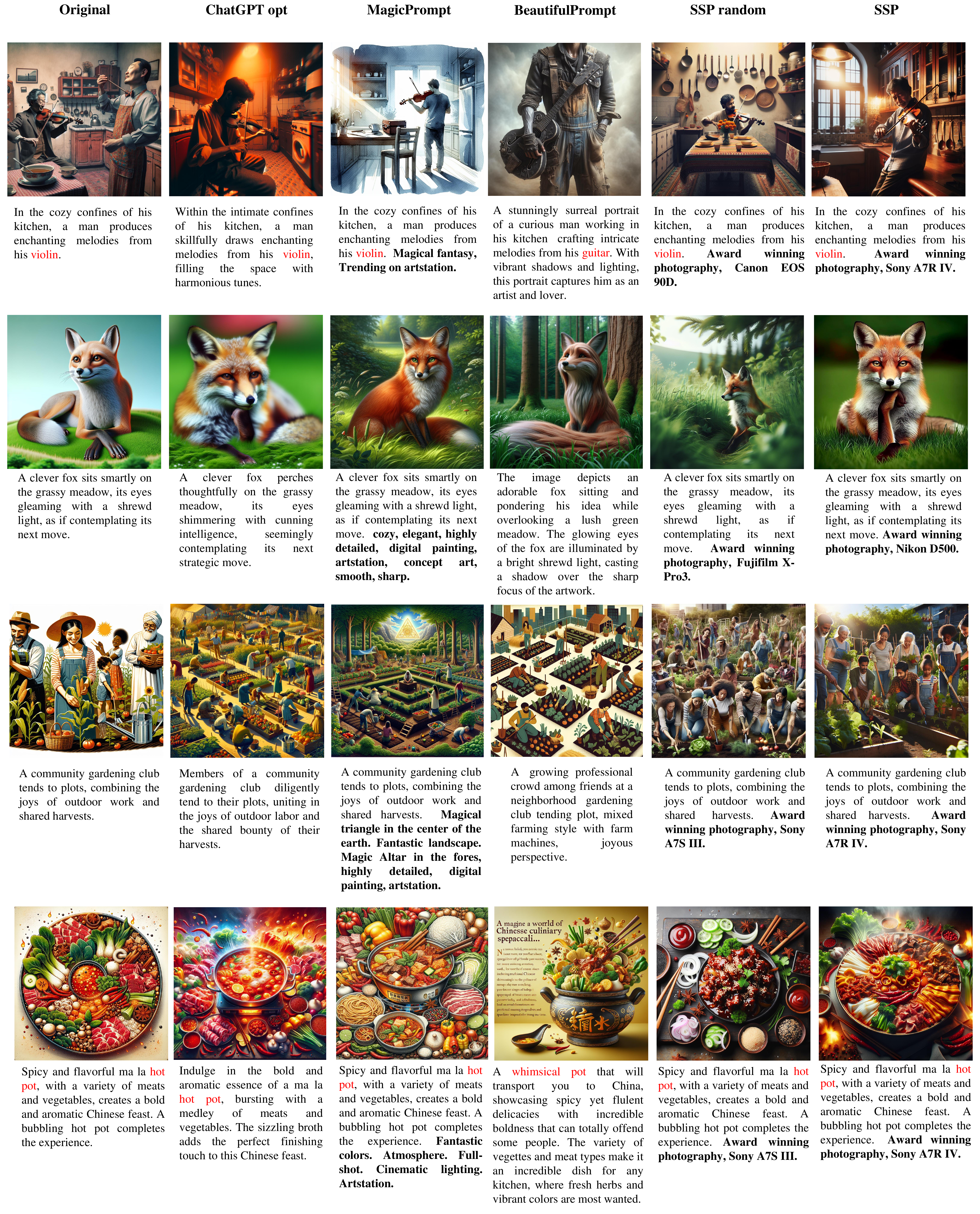}
	\caption{Examples of prompts and images generated by different methods. Red font denotes errors or omissions from other methods. Bold text indicates additions directly after the original prompt without altering it.}
	\label{tu2}
\end{figure*}

Some examples of prompts and images generated by different methods are shown in Fig.~\ref{tu1}and Fig.~\ref{tu2}. From the red font in the two images, it can be observed that other methods alter the semantic meaning, omit content, or introduce additional information to the original prompts. For example, in Fig.~\ref{tu1}, the fourth image in the first line changes the original prompt's ``tranquil'' to ``extremely chaotic''. The original prompt includes both teacher and student, but in the fourth image of the second line in Fig.~\ref{tu1}, only the teacher is present, omitting the student. In the fourth image of the third line in Fig.~\ref{tu1}, ``people'' is added, whereas the original prompt makes no mention of ``people''.

\bibliographystyle{IEEEbib}
\bibliography{icme2023}

\end{document}